\documentclass[conference]{IEEEtran}
\IEEEoverridecommandlockouts
\usepackage{cite}
\usepackage{amsmath,amssymb,amsfonts}
\usepackage{algorithm}
\usepackage{algorithmic}
\usepackage{graphicx}
\usepackage{textcomp}
\usepackage{xcolor}
\usepackage{caption}
\usepackage{subcaption}
\usepackage{multirow}
\usepackage{tabularx}
\usepackage{array}

\begin{document}

\title{Human trajectory prediction using LSTM with Attention mechanism}

\author{\IEEEauthorblockN{1\textsuperscript{st} Amin Manafi Soltan Ahmadi}
\IEEEauthorblockA{\textit{Department of Electrical and Computer Engineering} \\
\textit{Isfahan University of Technology}\\
Isfahan 84156-83111, Iran \\
a.manafi@ec.iut.ac.ir}
\and
\IEEEauthorblockN{2\textsuperscript{nd} Samaneh Hoseini Semnani}
\IEEEauthorblockA{\textit{Department of Electrical and Computer Engineering} \\
\textit{Isfahan University of Technology}\\
Isfahan 84156-83111, Iran \\
samaneh.hoseini@iut.ac.ir}

}

\maketitle

\begin{abstract}
In this paper, we propose a human trajectory prediction model that combines a Long Short-Term Memory (LSTM) network with an attention mechanism. To do that, we use attention scores to determine which parts of the input data the model should focus on when making predictions. Attention scores are calculated for each input feature, with a higher score indicating the greater significance of that feature in predicting the output. Initially, these scores are determined for the target human's position, velocity, and their neighboring individuals' positions and velocities. By using attention scores, our model can prioritize the most relevant information in the input data and make more accurate predictions.
We extract attention scores from our attention mechanism and integrate them into the trajectory prediction module to predict humans' future trajectories.
To achieve this, we introduce a new neural layer that processes attention scores after extracting them and concatenates them with positional information. We evaluate our approach on the publicly available ETH and UCY datasets and measure its performance using the final displacement error (FDE) and average displacement error (ADE) metrics. We show that our modified algorithm performs better than the Social LSTM in predicting the future trajectory of pedestrians in crowded spaces. Specifically, our model achieves an improvement of 6.2\% in ADE and 6.3\% in FDE compared to the Social LSTM results in the literature.

\end{abstract}


\section{INTRODUCTION}
The ability to predict human trajectories in crowded spaces is a critical task that has significant applications in robotics, autonomous driving, and crowd management. The rise of smart cities and the increasing population density in urban areas have made this task even more crucial. Traditional trajectory prediction methods have relied on handcrafted features and heuristics to capture individual movements, but these methods often fail to account for the social context of pedestrians, which plays a crucial role in determining their trajectories. Recently, deep learning techniques, particularly Long Short-Term Memory (LSTM) networks, have been successfully used in human trajectory prediction. A number of papers have been published in this area in recent years, which are briefly reviewed below.

{\bfseries Classic methods} that have been used for human trajectory prediction include rule-based methods, Bayesian methods, and hybrid methods. Rule-based methods use heuristics and predefined rules to predict pedestrian trajectories\cite{force_ped}. Bayesian methods use probability theory and Bayes' theorem to estimate the posterior probability of pedestrian trajectories based on observed data\cite{bayesian}. Hybrid methods combine multiple approaches, such as combining a rule-based method with a machine learning method, to improve prediction accuracy\cite{hybrid}. These methods have their strengths and weaknesses, but machine learning-based methods, especially deep learning-based methods, have shown significant improvements in trajectory prediction accuracy and have become the state-of-the-art methods in recent years.

To overcome the limitations of traditional methods, deep learning approaches have emerged as promising solutions. These methods can automatically learn representations from data and capture complex interactions between individuals. One popular approach in this area is the social LSTM, which extends the standard LSTM architecture to incorporate social interactions between individuals. While social LSTMs have shown remarkable performance in predicting human trajectories in crowded spaces, they still suffer from limitations such as over-reliance on past observations and a lack of attention mechanisms\cite{social_lstm}.

{\bfseries Social-LSTM-based methods} use Long Short-Term Memory (LSTM) networks to model the social interactions between pedestrians and predict their future trajectories\cite{social_lstm,social_gan,trajectron,social_attention,colab_recom,contex_aware}. These methods have shown promising results in improving trajectory prediction accuracy, especially in crowded scenarios where social interactions play a crucial role. Some examples of these methods include Social LSTM\cite{social_lstm}: uses an LSTM-based model to capture the social interactions among pedestrians and predict their future trajectories. Social GAN\cite{social_gan}: combines a generator network and a discriminator network to generate plausible future trajectories for pedestrians in a social setting. Trajectron++\cite{trajectron}: builds upon the Social GAN approach by using a graph-based representation to model the interactions between pedestrians and the environment. Social Attentional LSTM\cite{social_attention}: extends the Social LSTM approach by introducing an attention mechanism to selectively focus on the most relevant pedestrians in a scene. Collaborative LSTM for Human Trajectory Prediction\cite{social_lstm, colab_recom}: models the interactions between pedestrians as a series of pairwise interactions and uses a collaborative LSTM to predict their future trajectories. Context-Aware Social LSTM for Pedestrian Trajectory Prediction\cite{contex_aware}: incorporates contextual information such as weather and time of day into the Social LSTM model to improve trajectory prediction accuracy.

{\bfseries Graph-based methods} model the spatial relationships between pedestrians using graph structures and use them to predict their future trajectories\cite{gcnt,stgcnn,stgcnn,graph_attention,spatial_temporal_graph,heterogeneous_graph}. They represent the scene as a graph, where nodes are pedestrians and edges represent the spatial proximity or social interactions. The graph structure allows for flexible modeling of the interactions between pedestrians and capturing the global context. Some examples of these methods include Graph Convolutional Networks (GCN)\cite{gcnt,stgcnn}: a method that uses GCN to model the interactions between agents in a social network and predict their trajectories.
Social Graph Convolutional Networks (SGCN)\cite{stgcnn}: a method that extends GCN to include social relationships between pedestrians and improve the accuracy of trajectory prediction. Graph Attention Networks (GAT)\cite{graph_attention}: a method that uses attention mechanisms to selectively focus on certain nodes in the graph and improve the accuracy of trajectory prediction. Spatial-Temporal Graph Convolutional Networks (ST-GCN)\cite{spatial_temporal_graph}: a method that extends GCN to model the spatial-temporal dynamics of pedestrian interactions and improve the accuracy of trajectory prediction. Heterogeneous Graph Neural Networks (HGN)\cite{heterogeneous_graph}: a method that uses heterogeneous graphs to model the different types of interactions between pedestrians and improve the accuracy of trajectory prediction.
Social-STGCNN\cite{stgcnn}: a method that combines social interaction and spatial-temporal information using a spatio-temporal graph convolutional neural network to improve the accuracy of trajectory prediction.

{\bfseries Attention-based methods} use attention mechanisms to selectively focus on certain features or pedestrians in the input data and improve the accuracy of trajectory prediction. These methods assign different weights to different components of the input data, allowing the model to attend to the most relevant information for trajectory prediction. Some examples of attention-based methods include Attentional Pedestrian Prediction\cite{context_intraction}, Crowd Attentional Networks\cite{context_intraction}, and Attention-based Social GAN\cite{sophie}. These methods have shown promising results in improving the accuracy of pedestrian trajectory prediction. Attentional Pedestrian Prediction\cite{context_intraction}: This method uses an attention mechanism to predict pedestrian trajectories by selectively attending to certain past trajectory segments. Crowd Attentional Networks\cite{context_intraction}: this method uses a multi-head attention mechanism to capture the interactions between different individuals in a crowd and predict their future trajectories. Attention-based Social GAN\cite{sophie}: this method combines a social GAN with an attention mechanism to generate more realistic future trajectories by focusing on the relevant individuals in the scene. Attention-based Ensemble of LSTM Networks for 3D Human Trajectory Prediction\cite{threed_attention}: this method uses an ensemble of LSTM networks with an attention mechanism to predict 3D human trajectories by selectively focusing on relevant parts of the input data. Trajectory Prediction of Many Agents with Multimodal Attention and Graph Embedding\cite{bigat}: This method uses a graph embedding technique and a multimodal attention mechanism to predict the trajectories of multiple agents in a scene.

{\bfseries Reinforcement learning-based methods} use a trial-and-error approach to optimize the trajectory prediction model by maximizing a reward function. The model learns to make decisions based on the environment and the reward it receives for each action. Some examples of these methods include Deep Reinforcement Learning for Pedestrian Trajectory Prediction\cite{social_aware_rl} and Crowd-Robot Interaction\cite{crowd_robot}. These methods have shown promising results in improving the accuracy of trajectory prediction, especially in complex and dynamic environments where traditional methods may struggle. However, they also require a large amount of training data and can be computationally expensive. Some examples of these methods include"Deep Reinforcement Learning for Pedestrian Trajectory Prediction"\cite{social_aware_rl}: uses deep reinforcement learning to predict pedestrian trajectories in crowded scenes by incorporating social and physical constraints. "Crowd-Robot Interaction"\cite{crowd_robot}: proposes a reinforcement learning-based method for robots to navigate through crowds of pedestrians while minimizing the impact on their movements. "Socially Aware Motion Planning with Deep Reinforcement Learning"\cite{social_aware_rl}: uses a deep reinforcement learning approach for socially aware motion planning of autonomous vehicles. The model uses a value-based method with a social-aware reward function.

To address these issues, we propose a novel approach that combines a deep learning network with LSTM blocks and a social LSTM block, along with attention-based deep reinforcement learning algorithms. Our approach leverages the social context of pedestrians and generates attention scores for each individual in the scene, allowing us to focus on the most relevant information for trajectory prediction. We extract attention scores using reinforcement learning to optimize the scores while maximizing the prediction accuracy. However, our prediction model is based on a deep learning network with LSTM blocks and a social LSTM block.
We evaluated our approach on several benchmark datasets, including ETH\cite{eth}, and UCY\cite{ucy}, and demonstrated its superior performance compared to state-of-the-art methods. Our model is better than previous methods in terms of prediction accuracy and collision avoidance. Our work has significant implications for the development of intelligent systems that can safely and effectively navigate crowded environments. By accurately predicting human trajectories, our approach can enable robots and autonomous vehicles to navigate in crowded spaces with confidence and improve the safety and efficiency of crowd management systems.

This paper is organized as follows. In Section II, we provide background information on the research in the field of human trajectory prediction. We discuss the two base approaches\cite{social_lstm,crowd_robot} that form the foundation of our work, highlighting their contributions and limitations. Section III presents our proposed approach, called Attention-Social-LSTM, which is based on the integration of attention mechanism into the Social-LSTM framework. We explain the architecture of our model in detail, emphasizing on how it addresses the limitations of the existing methods.
In Section IV, we describe the experimental setup and methodology used to evaluate the performance of our proposed approach. We provide details on the datasets used, evaluation metrics employed, and any preprocessing or parameter tuning performed. The results of our experiments are presented in Section IV where we analyze and discuss the performance of our model compared to the baseline method (Social-LSTM). We show the effectiveness and reliability of our approach through quantitative metrics such as Average Displacement Error (ADE) and Final Displacement Error (FDE)\cite{eth}.
Finally, in Section V, we present conclusions based on the findings of our study. We summarize the main contributions of our work, highlight its significance in improving trajectory prediction accuracy, and discuss the implications for real-world applications. Additionally, we suggest potential areas for future research and development to further enhance the performance of human trajectory prediction models.

\section{BACKGROUND}

Human trajectory prediction is another important research area that has many applications, including robotics, autonomous vehicles, and surveillance systems. Predicting the future trajectories of humans is challenging, especially in crowded environments where the interactions between individuals are complex and unpredictable.

\subsection{Crowd-Robot Interaction: Crowd-aware Robot Navigation with Attention-based Deep Reinforcement Learning}

\begin{figure}
  \centering
  \includegraphics[scale=0.22]{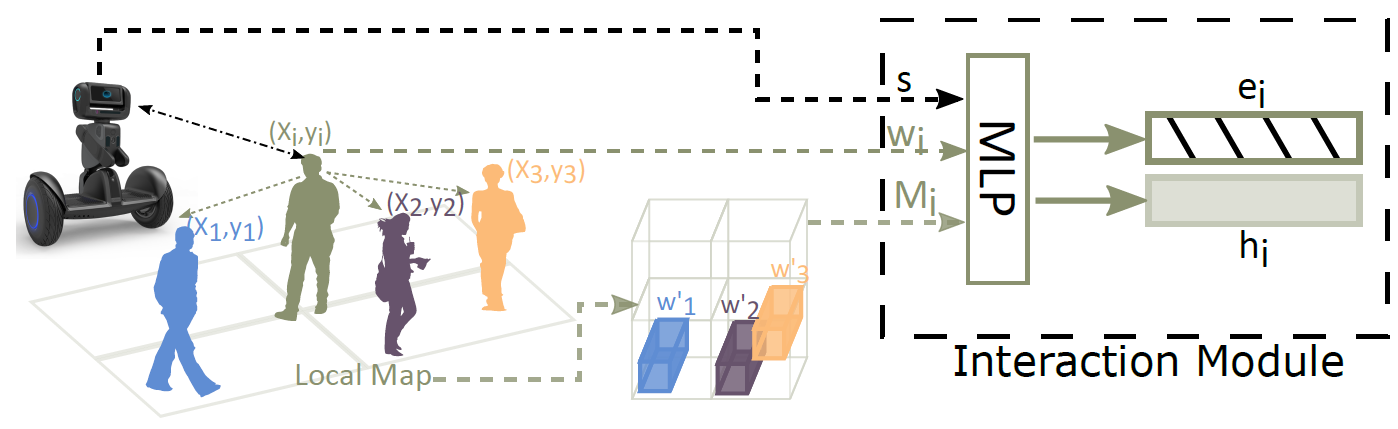}
  \caption{Interaction Module Overview. The interaction module incorporates information from both humans and the robot, represented by the local map, and passes it through a multi-layer perceptron (MLP). This MLP extracts pairwise interaction features between the robot and each human present in the scene, enabling effective modeling of their interactions\cite{crowd_robot}.}
  \label{fig.attention1}
\end{figure}

\begin{figure}
  \centering
  \includegraphics[scale=0.22]{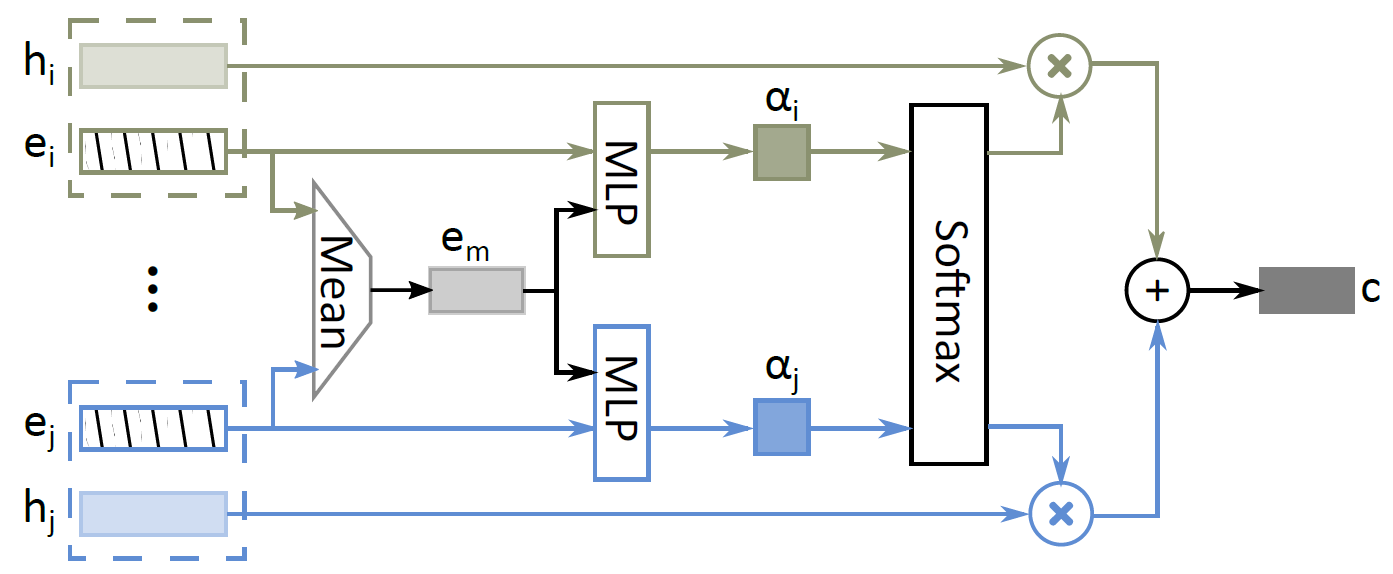}
  \caption{Pooling Module Overview. The attention features extracted in the interaction module are forwarded to the pooling module. By employing a multi-layer perceptron with the input of the individual embedding vector and the mean embedding vector, attention scores are computed for each human. Subsequently, a weighted sum of the pairwise interactions is calculated using softmax activation to obtain the final representation\cite{crowd_robot}.}
  \label{fig.attention2}
\end{figure}

Crowd-Robot Interaction is an important research topic that has gained a lot of attention in recent years due to the increasing demand for robots to navigate safely and efficiently in crowded environments. One of the major challenges in this area is to enable robots to interact with humans in a socially acceptable manner while avoiding collisions and achieving their objectives.

In\cite{crowd_robot} Changan Chen, et al. propose a novel approach for robot navigation in crowded environments. The main focus of the paper is to address the challenge of safely and efficiently navigating robots through dense crowds while ensuring minimal disruption to the pedestrians' movements. To achieve this, they combine deep reinforcement learning with attention mechanisms. The attention mechanism computes attention scores for both the target human and their neighbors. This aids the robot in trajectory planning by indicating the degree of attention that should be directed towards each neighbor during the planning process, thereby minimizing the risk of collisions. By leveraging the attention-based deep reinforcement learning framework, the robot learns to make context-aware navigation decisions that consider the dynamics and interactions of the crowd. The experimental results demonstrate that the proposed method significantly improves the robot's navigation performance, leading to more efficient and socially aware interactions with the crowd.

The key contribution of the paper is the use of attention-based deep reinforcement learning to improve the performance of the local planner. The attention mechanism allows the robot to focus on the most relevant parts of the environment and take into account the behavior of nearby pedestrians when generating trajectories.

In this paper, the authors address the challenge of modeling interactions among humans in a computationally efficient manner. They introduce a pairwise interaction module that captures the Human-Robot interaction while using local maps to represent the interactions between humans. To encode the presence and velocities of neighboring humans, a map tensor is constructed for each individual, centered around them. The map tensor, referred to as the local map, is constructed as a grid and contains information about the relative positions of other humans. By combining the state of the human, the local map, and the state of the robot, an embedding vector is created using a multi-layer perceptron (MLP) with ReLU activations.
This embedding vector is then fed into another MLP to obtain the pairwise interaction feature between the robot and each human. This interaction feature is obtained through a fully-connected layer with ReLU activations which is shown in Fig. \ref{fig.attention1}.

Then the pooling module is introduced to address the challenge of handling an arbitrary number of inputs (surrounding humans) and producing a fixed-size output representation. The purpose of the pooling module is to use neighbors' information in the next time step to enhance trajectory prediction. To achieve this, we store the output of LSTM blocks of neighbors of the target human in the pooling layer. In the subsequent time step, these stored outputs are utilized to predict the future trajectory more effectively. The pooling module aims to learn the relative importance of each neighbor and the collective impact of the crowd in a data-driven fashion. By leveraging the self-attention mechanism, the pooling module computes an attention score for each interaction embedding, capturing the importance of the corresponding neighbor. The attention scores are obtained by transforming the interaction embeddings and applying a social attentive pooling approach. The final representation of the crowd is then obtained through a weighted linear combination of all the pairwise interactions, using the attention scores as weights which is shown in Fig. \ref{fig.attention2}\cite{crowd_robot}.

\subsection{Social LSTM: Human Trajectory Prediction in Crowded Spaces}

\begin{figure}
  \centering
  \includegraphics[scale=0.45]{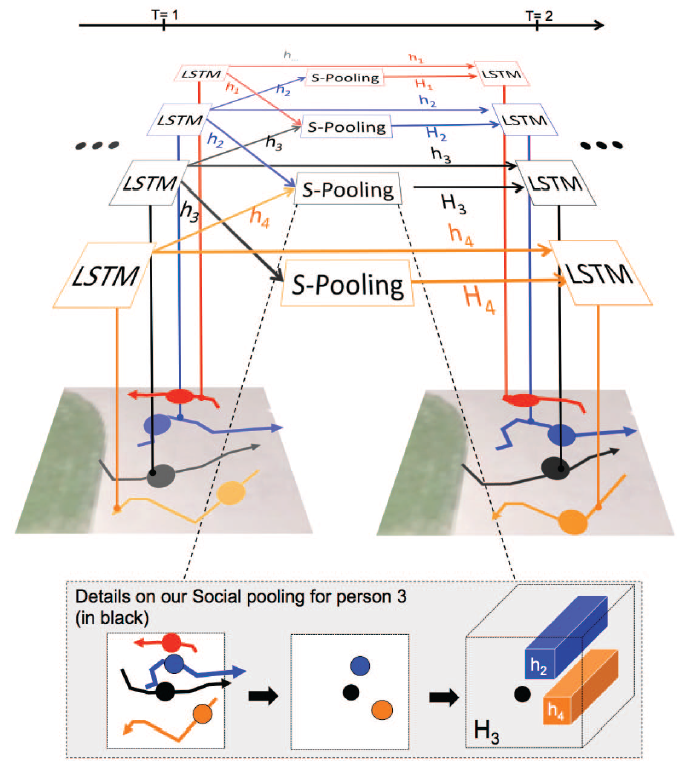}
  \caption{Overview of the social-LSTM method. The social-LSTM approach involves the transmission of previous positions to LSTM blocks. Subsequently, the output of LSTMs for each individual is connected using a pooling layer to facilitate the exchange of neighbor information, as illustrated in the lower part of the figure (highlighting the details for one person). Finally, these combined features, along with the LSTM's internal state, are fed back into the LSTM blocks to predict the next steps\cite{social_lstm}.}
  \label{fig.s-lstm1}
\end{figure}

Alexandre Alahi et al. proposed a deep learning-based approach called Social LSTM to predict the future trajectories of humans in crowded spaces. The proposed method takes into account the social interactions between individuals and learns to predict their future trajectories based on their past movements and the movements of their neighbors.

The key contribution of the paper is the use of a social pooling mechanism that aggregates the information from the past movements of the neighbors and incorporates it into the prediction process. The social pooling mechanism allows the model to capture the social dynamics of the environment and make accurate predictions even in highly crowded scenarios.The method presented in Alahi's work is shown in Fig. \ref{fig.s-lstm1}\cite{social_lstm}.

\section{Approach}

\begin{figure}
  \centering
  \includegraphics[scale=0.3]{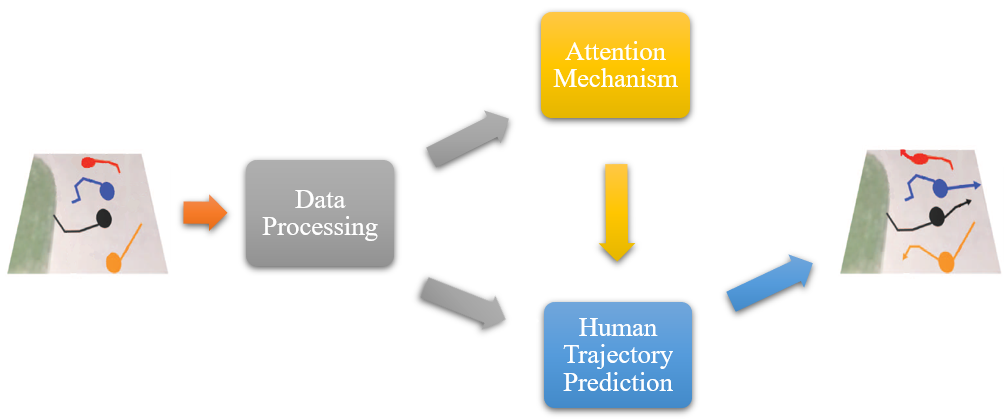}
  \caption{Block diagram of Attention Social LSTM method.}
  \label{fig.diagram}
\end{figure}

This section presents our innovative methodology for human trajectory prediction using LSTM with an attention mechanism. It encompasses three key components: data processing, attention mechanism, and human trajectory prediction. We first discuss the data processing techniques employed to preprocess the input trajectory data and create a suitable representation. Next, we introduce the attention mechanism, a crucial aspect of our approach, which enables the model to effectively capture contextual information. Finally, we outline the human trajectory prediction process, leveraging LSTM to learn temporal dependencies and generate accurate predictions. By systematically addressing these components, our approach aims to advance the field of human trajectory prediction through enhanced accuracy and contextual understanding. As illustrated in Fig. \ref{fig.diagram}, our approach follows a sequential process. Initially, we gather information about humans from the environment. Subsequently, we preprocess this data to prepare it for input into our networks. The processed data is then directed to the attention mechanism, from which we extract attention scores. These attention scores provide insights into the significance of each human's contextual information. Lastly, both the processed data and attention scores are fed into our predictor network, enabling us to predict the future trajectory for each individual accurately. This holistic process showcases how our methodology integrates attention mechanism and prediction network to enhance trajectory prediction outcomes.

\subsection{Data Processing}

\begin{figure}
  \centering
  \includegraphics[scale=0.8]{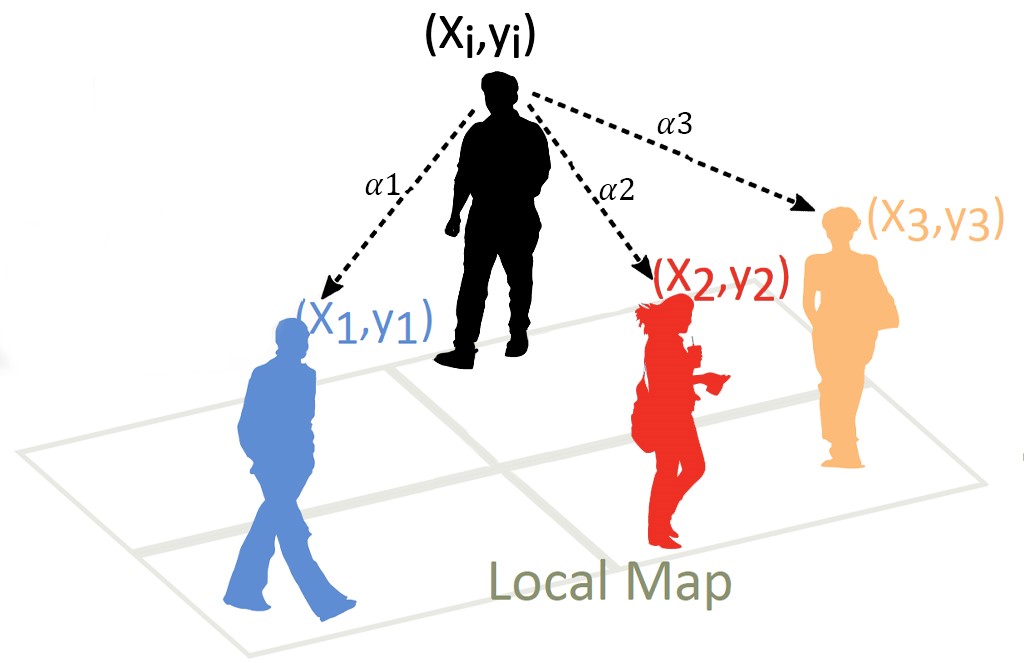}
  \caption{Attention Module: In our attention algorithm, we modify the interaction term from robot-human to human-human. Within the local map, our focus is on predicting the future trajectory of the i-th person, and therefore,in this example we extract three attention scores corresponding to three neighbors.}
  \label{fig.approach1}
\end{figure}

In our paper, we utilize the UCY and ETH datasets for human trajectory prediction. These datasets provide information such as the frame number, human number, and the corresponding x and y positions of each human. However, as the attention mechanism algorithm requires additional information, we perform slight modifications to the dataset during the preprocessing stage.

In particular, we enhance the dataset by including speed information when submitting data to extract attention scores. To get the velocity in the x direction, we use \ref{velocityx}, where $x_{i}^{t}$ represents the spatial coordinates of x for the $ith$ person at time $t$, $x_{i}^{t}$ represents the spatial coordinates of x for the $i th$ person at time $t-1$ and $\Delta t$ is the time spent to move between two spatial points, and also to get the velocity in the y direction from the \ref{velocityy}, we use that $y_{i}$ indicates the location coordinates in the y direction for the $i th$ person. Once the velocity data has been incorporated, the modified data set is submitted for execution in the attention mechanism algorithm.

In the next part, because the movement path data is a time series and the future movement path is dependent on the future movement path, we can use this sequence in LSTM blocks to predict the future movement path, so as in\cite{social_lstm}, data We have created a time series of 20 frames, the first 8 frames represent the previous movement path of each human, and the next 12 frames are used for prediction and evaluation.

\begin{equation}\label{velocityx}
  v_{xi} = \frac{x_{i}^{t} - x_{i}^{t-1}}{\Delta t}
\end{equation}
\begin{equation}\label{velocityy}
  v_{yi} = \frac{y_{i}^{t} - y_{i}^{t-1}}{\Delta t}
\end{equation}

\subsection{Attention Mechanism}

The attention mechanism plays a crucial role in our approach, aiming to determine the extent to which each individual should focus on its neighboring humans when predicting its future trajectories. By assigning attention scores to each human with respect to its neighbors, we enhance the prediction process and improve its accuracy.

In contrast to previous approaches that relied on the robot's goal position for extracting attention scores during motion planning, our focus is on prediction tasks where the goal position is unknown. To accommodate this, we modify the attention mechanism by eliminating the explicit inclusion of the goal position information. In the following, due to the use of this mechanism for extracting human interactions, we change this network from robot-human mode to human-human mode and instead of sending information about the robot, we send information about the target human to the network. By doing this, we obtain human-human attention scores that represent the relative importance of each neighbor in influencing the future path of the target human which is visualized in Fig. \ref{fig.approach1}.

In our modified attention mechanism, the attention scores are computed by considering the interactions and dynamics among humans. By analyzing the positions, velocities, and other relevant features of neighboring humans, we extract valuable cues about their influence on the target individual's trajectory. These attention scores guide the subsequent stages of our prediction model, enabling it to generate more accurate and refined future trajectory predictions.

\subsection{Human Trajectory Prediction}

\begin{figure}
  \centering
  \includegraphics[scale=0.12]{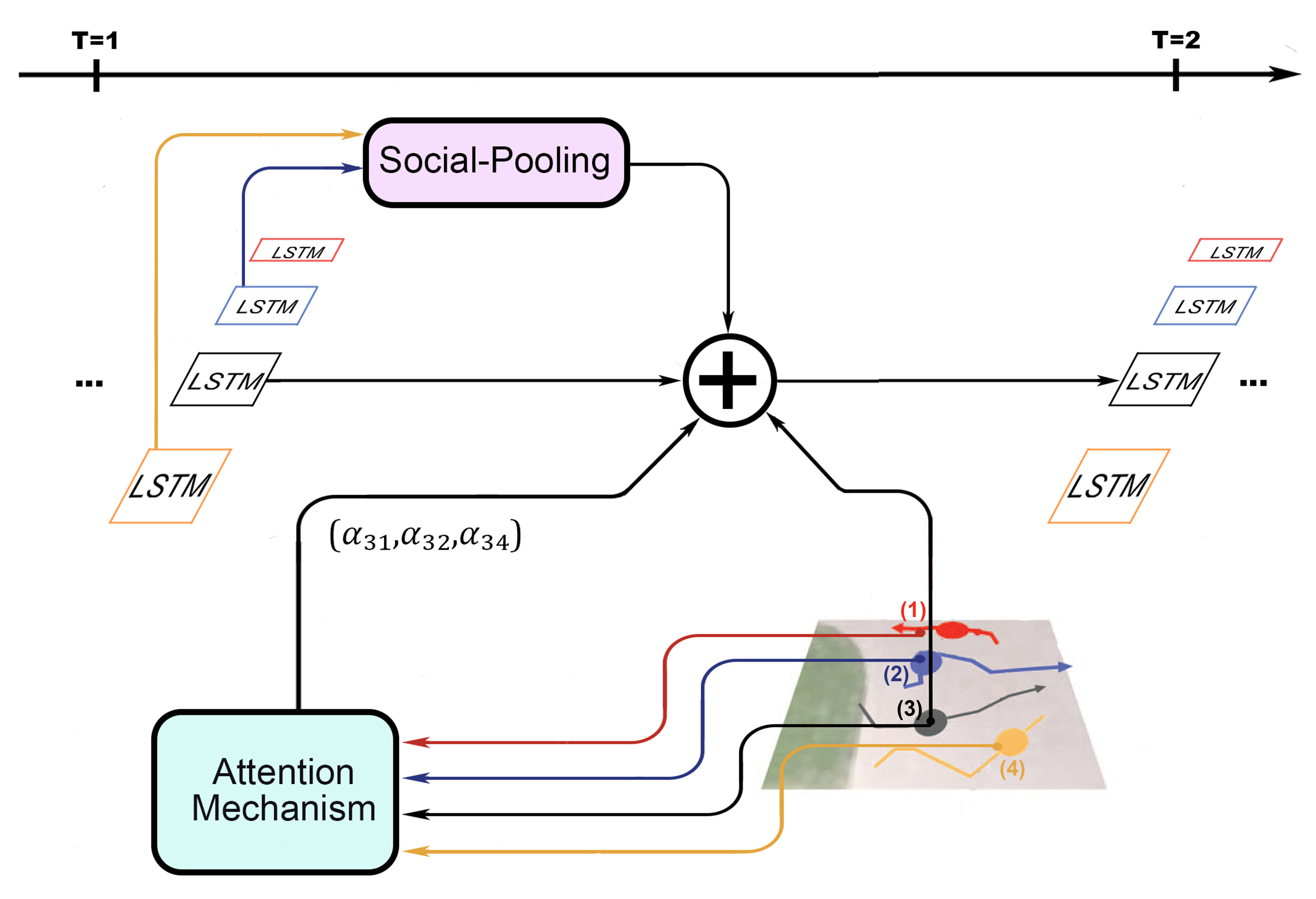}
  \caption{Attention-Social-LSTM Module: The attention mechanism captures attention scores, denoted as $\alpha_{31}$, $\alpha_{32}$, and $\alpha_{34}$, for all pedestrians in the scene, with a specific focus on the target individual (represented as a black human on the map). These attention scores are obtained using an attention mechanism from the local map.
    Additionally, the social-pooling component stores the previous information of the target's neighbors. To predict the future trajectory, we concatenate the attention scores, social-pooling information, target's position, and the previous time step's information from the LSTM block. This concatenated information is then fed as input to the LSTM block, enabling accurate future trajectory prediction.}
  \label{fig.approach2}
\end{figure}

In the prediction module, we employ LSTM blocks for each individual to generate accurate trajectory predictions. After gathering relevant information from the Data Processing and Attention Mechanism, the positional data and attention scores are processed using a linear layer with ReLU activation function. These processed inputs have the same size and align with the dimensions of the social-pooling module.

To incorporate the information from neighboring individuals, we utilize the previous time step's data of these neighbors in the social-pooling module. By including the neighbors' information, our model benefits from the collective behavior of the surrounding individuals.

In the trajectory prediction part, we concatenate the target individual's previous predictions (LSTM outputs) who we want to predict its future trajectory, with the processed positional data, attention scores, and social-pooling module outputs. This concatenated input is then fed into the LSTM block for predicting the future trajectory. The overall architecture of our prediction module is depicted in Fig. \ref{fig.approach2}.

\begin{algorithm}
\caption{Attention-Social-LSTM}
\label{alg1}
\begin{algorithmic}[1]
\STATE \textbf{Input:} $Data=\sum_{i=1}^{n}Data_{i}, Data_{i}=(frame\_fum, ped\_ID, x, y)$
\STATE \textbf{Output:} Prediction of humans' future trajectory (12 frames) in the scene
\FOR{\textbf{each} $ human_{i} $}
\STATE $social\_pooling \leftarrow [...]$
\FOR{\textbf{each} $ neighbor_{i,j}$ }
\STATE $social\_pooling_{i} \leftarrow prediction_{j}^{t-1} + social\_pooling_{i}$
\STATE $\alpha_{i,j} \leftarrow attention\_mechanism(Data_{j})$
\ENDFOR
\STATE $prediction_{i} \leftarrow LSTM ( Data_{i} + social\_pooling_{i} + \alpha_{i} )$
\ENDFOR
\end{algorithmic}
\end{algorithm}

\section{Experiments}

\subsection{Setup}
In this section, we provide a comprehensive overview of the experimental setup used to evaluate the performance of our proposed human trajectory prediction approach using LSTM with attention mechanism. The following details are provided:

Local Map and Position Calculation:
We assume the human for whom we want to predict the future trajectory is positioned at the center of the local map with coordinates (x=0, y=0). The positions of other individuals are set relative to this central human. Velocity information is calculated for each human and its neighbors in the UCY and ETH datasets, which is used to initialize the network.

Attention Mechanism Modification:
The dataset used for the attention mechanism does not contain goal positions. So the goal positions are removed from the attention mechanism's network, as they are not available in the prediction task and the state of robot-human interaction is changed to human-human interaction.

Local Map Configuration:
The local map is a 4x4 grid centered around each human, with each cell having a side length of 1m. This grid-based representation allows for capturing spatial relationships between individuals effectively.

Network Architecture:
The attention mechanism network comprises several hidden units with ReLU activations. The embedding function consists of two fully-connected layers with dimensions (150, 100), followed by a fully-connected layer with ReLU nonlinearity of dimension (100, 50). The MLP (multi-layer perceptron) component has a dimension of (100, 100) with ReLU activations. The entire network is implemented in PyTorch.

Training Parameters:
The attention mechanism network is trained using the Adam optimizer with a batch size of 100. Imitation learning is performed using 3,000 episodes of demonstrations collected using ORCA\cite{orca1,orca2}, and the policy is trained for 50 epochs with a learning rate of 0.01. Reinforcement learning utilizes a learning rate of 0.001 and a discount factor (gamma) of 0.9. The exploration rate of the e-greedy policy linearly decays from 0.5 to 0.1 over the first 5,000 episodes and remains constant at 0.1 for the remaining 5,000 episodes. The RL training process takes approximately 20 hours on an Intel® Core™ i7-9700K CPU @ 3.60GHz × 8.

Social LSTM Trajectory Prediction:
For the Social LSTM model, we employ an embedding dimension of 64 for the spatial coordinates. The position data and attention scores are processed using neural networks (nn) with dimensions of (2, 64) and (1, 64), respectively, both with ReLU activation functions. The social tensor is processed using an MLP with dimensions (1024, 64) and a ReLU activation function. Dropout layers are applied to all the aforementioned layers. The concatenated data is then fed into an LSTM block with dimensions (3 * 64, 128) and subsequently to an output layer with dimensions (128, 5) and a linear activation function. The spatial pooling size N is set to 32, and an 8x8 sum pooling window size without overlaps is used. The LSTM models have a fixed hidden state dimension of 128. An embedding layer with ReLU non-linearity is applied to the pooled hidden state features before calculating the hidden state tensor. The hyperparameters are determined through cross-validation on a synthetic dataset generated using the social forces model. Training is performed with a learning rate of 0.003 and RMSprop optimization. The Social-LSTM model is trained using a NVIDIA Corporation TU102 [GeForce RTX 2080 Ti Rev. A] graphics card and an Intel® Core™ i7-9700K CPU @ 3.60GHz × 8.

%algorithm description
As outlined in Algorithm \ref{alg1}, we utilize input data $(Data)$ containing positional coordinates (x, y), pedestrian ID, and frame number. The algorithm predicts the future trajectory (12 frames) for each individual $human_{i}$ in the scene.
For each $human_{i}$, we construct a social pooling $(social\_pooling_{i})$ by aggregating the previous trajectory predictions of their neighboring pedestrians. This social tensor captures the influence of neighbors on $human_{i}$'s trajectory prediction (line 6).
We calculate the attention score $(\alpha_{i,j})$ for each $human_{i}$ and their $neighbor_{j}$, quantifying the relevance of each neighbor to $human_{i}$'s trajectory (line 7).
Once we have gathered the social tensors and attention scores for $human_{i}$ and their neighbors, we feed the data to an LSTM block. The LSTM takes as input the individual data of $human_{i}$ $(Data_{i})$, the $social\_pooling_{i}$, and the attention scores $(\alpha_{i})$. These inputs are concatenated using the '+' operator (line 9), enabling the LSTM to effectively incorporate the contextual information.

With this detailed experimental setup, we aim to evaluate the effectiveness of our proposed approach for human trajectory prediction using LSTM with attention mechanism. In the following section, we present the results obtained from these experiments.

\subsection{Result}

In this section, we present the results of our method's evaluation using various metrics, including the Average Displacement Error (ADE) and Final Displacement Error (FDE). These metrics serve as standard measures to assess the accuracy of trajectory prediction models\cite{social_lstm,eth,trajectron,social_attention}. Our method was evaluated on the test set using these parameters to evaluate its performance.

\subsubsection{Evaluation Metrics}
The ADE measures the average Euclidean distance between the predicted and ground truth trajectories at each time step. It provides insights into the overall accuracy of trajectory predictions. On the other hand, the FDE quantifies the Euclidean distance between the final predicted position and the actual final position of the target trajectory. This metric captures the model's ability to accurately predict the destination point.

\subsubsection{Plots Analysis}

\begin{figure}
  \centering
  \begin{subfigure}{0.5\textwidth}
  \includegraphics[scale=0.6]{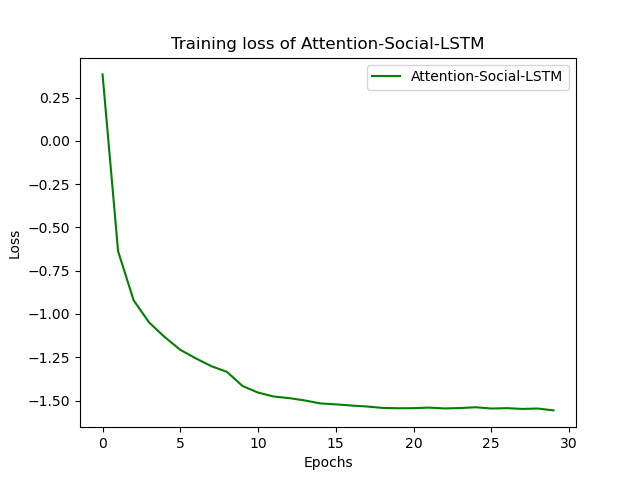}
  \caption{Training Loss for Attention-Social-LSTM}
  \label{fig.training-loss}
  \end{subfigure}
  \begin{subfigure}{0.5\textwidth}
  \includegraphics[scale=0.6]{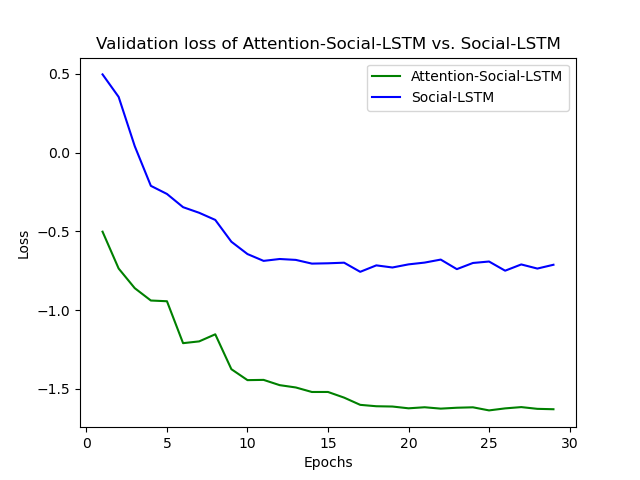}
  \caption{Validation Loss Comparison: Attention-Social-LSTM vs. Social-LSTM}
  \label{fig.both-loss}
  \end{subfigure}
  \caption{(a) Training Loss: The plot demonstrates the decreasing trend of the training loss over 30 epochs, indicating the convergence of our Attention-Social-LSTM model.
(b) Validation Loss: A comparison between our method and Social-LSTM on the validation dataset, highlighting our method's consistently lower loss, signifying its superior performance.}
  \label{fig.losses}
\end{figure}

Fig. \ref{fig.training-loss} shows the training loss of our method over the course of 30 epochs. The y-axis represents the loss value, while the x-axis denotes the number of epochs. As observed, the training loss steadily decreases, indicating that our model effectively converges and learns from the training data. This plot demonstrates the successful training process of our Attention-Social-LSTM algorithm and the Fig. \ref{fig.both-loss} focuses on the validation loss and compares our method with the Social-LSTM approach. The y-axis represents the loss value, and the x-axis represents the epochs same as the previous plot. This plot clearly illustrates that our method consistently outperforms the Social-LSTM method in terms of loss on the validation set. Our method exhibits significantly lower loss values, highlighting its superior performance and robustness in capturing the underlying patterns in the data.

\begin{figure}
  \centering
  \begin{subfigure}{0.5\textwidth}
  \includegraphics[scale=0.6]{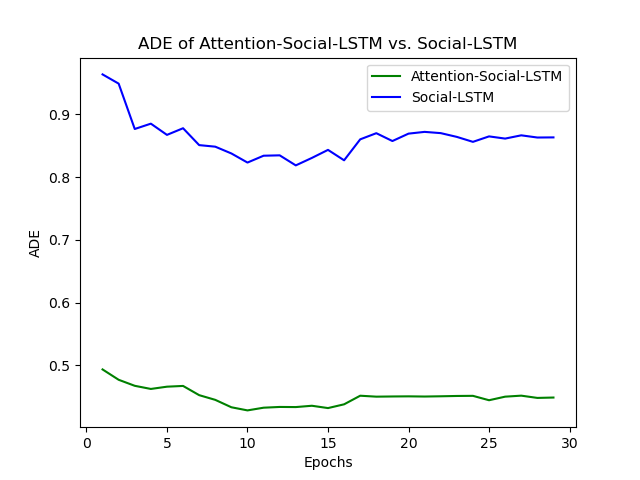}
  \caption{ADE Comparison: Attention-Social-LSTM vs. Social-LSTM}
  \label{fig.mean-error}
  \end{subfigure}
  \begin{subfigure}{0.5\textwidth}
  \includegraphics[scale=0.6]{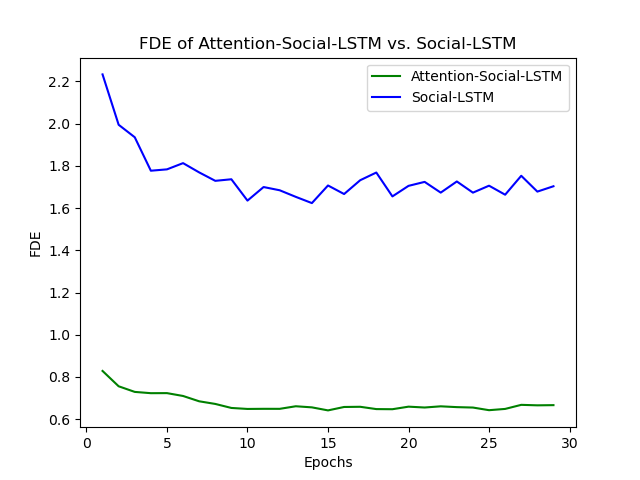}
  \caption{FDE Comparison: Attention-Social-LSTM vs. Social-LSTM}
  \label{fig.final-error}
  \end{subfigure}
  \caption{(a) ADE Comparison: The plot illustrates the comparison of Average Displacement Error (ADE) between our Attention-Social-LSTM method and Social-LSTM on the validation dataset. Our method consistently exhibits lower ADE, showcasing its improved accuracy in trajectory prediction.
(b) FDE Comparison: The plot compares the Final Displacement Error (FDE) between our method and Social-LSTM on the validation dataset. Our method consistently outperforms Social-LSTM, demonstrating its enhanced capability in accurately estimating the final destination of human trajectories.}
  \label{fig.errors}
\end{figure}

Moving on to the error analysis, the Fig. \ref{fig.mean-error} compares the Average Displacement Error (ADE) between our method and the Social-LSTM approach. The y-axis represents the ADE, while the x-axis denotes the epochs. Throughout the validation dataset, our method consistently achieves lower Average Displacement Error (ADE) values compared to the Social-LSTM approach. This plot clearly demonstrates the enhanced accuracy of our approach in predicting human trajectories, as it exhibits significantly less error. Specifically, our method achieves an average reduction of approximately 0.4 in ADE compared to the Social-LSTM method. This substantial improvement in prediction accuracy highlights the effectiveness of our Attention-Social-LSTM algorithm in capturing the complexities of social interactions and producing more precise trajectory predictions and Fig. \ref{fig.final-error} specifically examines the Final Displacement Error (FDE) and provides a comparison between our method and the Social-LSTM approach. Our method consistently outperforms Social-LSTM in terms of FDE throughout the 30 epochs on the validation dataset. Notably, our approach achieves an average reduction of approximately 1.1 in FDE compared to the Social-LSTM method. This substantial improvement underscores the superior predictive capability of Attention-Social-LSTM algorithm in accurately estimating the final destination of human trajectories.

The comprehensive analysis of these four plots emphasizes the strong performance and superiority of Attention-Social-LSTM algorithm compared to the Social-LSTM method. The plots show the convergence of our model during training, the improved loss performance on the validation set, and the reduced errors in both ADE and FDE metrics. These findings provide compelling evidence of the effectiveness and reliability of our method in accurately predicting human trajectories.

\subsubsection{Test Results}

To provide a comprehensive evaluation of our method's performance, we present the Test Results table. The TABLE \ref{tab1}  summarizes the ADE and FDE metrics for both Social-LSTM and Attention-Social-LSTM across different datasets. In the ADE section, our method demonstrates lower errors in most datasets, indicating improved trajectory prediction accuracy. On average, our method achieves an ADE of 1.8822, outperforming Social-LSTM's ADE of 2.0067, resulting in a remarkable 6.2\% improvement.

Similarly, in the FDE section, our method consistently performs better than Social-LSTM. Our method achieves an average FDE of 3.2277, compared to Social-LSTM's FDE of 3.4321, indicating a substantial 6.3\% improvement in trajectory prediction accuracy.

The Test Results table clearly highlights the superior performance of our Attention-Social-LSTM method across a range of real-world datasets. The bolded cells in the table represent lower errors, further emphasizing the significant improvement in trajectory prediction accuracy achieved by our proposed method.

\begin{table}[ht]
\renewcommand{\arraystretch}{2}
\caption{Performance Comparison of Attention-Social-LSTM and Social-LSTM}
\label{tab1}
    \centering   
    \begin{tabular}{|c|c|c|c|}
      \hline
      \textbf{Metric} & \textbf{Dataset} & \textbf{Social-LSTM} & \textbf{Attention-Social-LSTM}  \\
      \hline
      \hline
      \multirow{7}{*}{\textbf{ADE}} & ETH & 3.8362  & \textbf{3.6030} \\
      \cline{2-4}
      & HOTEL & \textbf{2.3760} & 2.4147 \\
      \cline{2-4}
      & UNIV1 & 1.3438 & \textbf{1.3125} \\
      \cline{2-4}
      & UNIV3 & 1.4145 & \textbf{1.3320} \\
      \cline{2-4}
      & ZARA1 & 1.4852 & \textbf{1.2586} \\
      \cline{2-4}
      & ZARA2 & 1.5847 & \textbf{1.3729} \\
      \cline{2-4}
      & Average & 2.0067 & \textbf{1.8822} \\
      \hline
      \hline
      \multirow{6}{*}{\textbf{FDE}} & ETH & 6.8811  & \textbf{6.2144} \\
      \cline{2-4}
      & HOTEL & \textbf{4.0892} & 4.2341 \\
      \cline{2-4}
      & UNIV1 & 2.2909 & \textbf{2.2941} \\
      \cline{2-4}
      & UNIV3 & 2.4191 & \textbf{2.1943} \\
      \cline{2-4}
      & ZARA1 & 2.4192 & \textbf{2.1858} \\
      \cline{2-4}
      & ZARA2 & 2.4935 & \textbf{2.2440} \\
      \cline{2-4}
      & Average & 3.4321 & \textbf{3.2277} \\
      \hline
    \end{tabular}
    
\end{table}

\section{Conclusions}
In this paper, we presented a novel approach for human trajectory prediction using the Attention-Social-LSTM algorithm. Our approach introduces the integration of attention mechanism, which captures social interactions, into the Social-LSTM framework to enhance prediction accuracy in crowded scenarios. We aimed to improve the accuracy of trajectory predictions, particularly in crowded scenarios where social interactions play a significant role. Through our extensive evaluation and analysis, we have obtained compelling results that demonstrate the effectiveness of our method.

First, we introduced two evaluation metrics, Average Displacement Error (ADE) and Final Displacement Error (FDE), to assess the performance of our method. ADE measures the average error in predicting future positions, while FDE evaluates the error in estimating the final destination of trajectories. These metrics provide a comprehensive understanding of the predictive capabilities of our model.

To validate the performance of our approach, we conducted a series of experiments and analysis. We plotted the training and validation loss curves, which showed the convergence of our model during training. The results clearly demonstrated that our method outperforms the baseline Social-LSTM approach, consistently achieving lower ADE and FDE values throughout the validation dataset. Specifically, our method exhibited around 0.4 less ADE error and approximately 1.1 less FDE error compared to Social-LSTM.

Furthermore, we presented a comprehensive table summarizing our test results on various datasets. The table highlighted the performance of our method in terms of ADE and FDE on each dataset, showing its superiority over the Social-LSTM approach. In most of the evaluated datasets, our method consistently achieved lower ADE and FDE errors, indicating its robustness and accuracy. Notably, our approach showed an average ADE error of 1.8822, outperforming Social-LSTM's average error of 2.0067, corresponding to a significant improvement of more than 6.2\%. Similarly, our method exhibited an average FDE error of 3.2277, surpassing Social-LSTM's average error of 3.4321 by more than 6.3\%.

In conclusion, our proposed Attention-Social-LSTM algorithm demonstrates remarkable effectiveness in predicting human trajectories, outperforming the baseline Social-LSTM method in terms of accuracy and error reduction. The consistent improvement across multiple evaluation metrics and datasets further establishes the robustness and generalizability of our approach. The findings of this study contribute to the field of trajectory prediction and hold significant implications for applications such as autonomous navigation, crowd management, and pedestrian safety.

Future work can explore additional enhancements to our model, such as incorporating contextual information or exploring advanced attention mechanisms. Furthermore, investigating the scalability and real-time performance of our method on larger and more complex datasets would be valuable for practical deployment.

Overall, our research presents a promising advancement in human trajectory prediction and sets the stage for further exploration and advancements in this domain.

\section*{Data Availability}
The data that support the findings of this study are available at https://github.com/quancore/social-lstm.

\section*{Conflict of Interest}
The authors declare that they have no conflict of interest.
\bibliographystyle{ieeetr}
\bibliography{refs}

\end{document}